# Transferable neural networks for enhanced sampling of protein dynamics

Mohammad M. Sultan[1], Hannah K. Wayment-Steele[1], & Vijay S. Pande[1†]

[1]Department of Chemistry, Stanford University, 318 Campus Drive, Stanford, California 94305, USA.

[†]pande@stanford.edu

**Abstract**
Variational auto-encoder frameworks have demonstrated success in reducing complex nonlinear dynamics in molecular simulation to a single non-linear embedding. In this work, we illustrate how this non-linear latent embedding can be used as a collective variable for enhanced sampling, and present a simple modification that allows us to rapidly perform sampling in multiple related systems. We first demonstrate our method is able to describe the effects of force field changes in capped alanine dipeptide after learning a model using AMBER99. We further provide a simple extension to variational dynamics encoders that allows the model to be trained in a more efficient manner on larger systems by encoding the outputs of a linear transformation using time-structure based independent component analysis (tICA). Using this technique, we show how such a model trained for one protein, the WW domain, can efficiently be transferred to perform enhanced sampling on a related mutant protein, the GTT mutation. This method shows promise for its ability to rapidly sample related systems using a *single* transferable collective variable and is generally applicable to sets of related simulations, enabling us to probe the effects of variation in increasingly large systems of biophysical interest.

**Introduction**
Efficient sampling of protein dynamics remains an unsolved problem in computational biophysics. Even with advances in GPU hardware, custom chips, and algorithms[1,2], most molecular dynamics (MD) simulation studies are limited to understanding the atomistic dynamics of one protein system at a time.

However, for MD to be predictive in guiding experiments, we require methods capable of describing the effects of perturbations to a system[3]. A perturbation can be very broadly defined, and could be a mutation to a protein sequence, a post-translational modification, ionic concentration, solvent type, protonation state, or chemical potential, or for better understanding simulation parameters, a change in force field (FF). For instance, we would like to predict via simulation how these perturbations affect protein dynamics, for instance, characterizing how a protein's folded state is stabilized or an intermediate is trapped.

Several analytical methods have been developed to combine information from simulation at multiple conditions to be able to make predictions about the system in different thermodynamic states (WHAM, bin-less WHAM, MBAR, DHAM, xTRAM, TRAM, etc.). However, there is also much information to be gained from an original simulation that can be leveraged to accelerate new simulations of related systems. If the phase space visited by a perturbed system is mostly unchanged from the original system, predicting these changes should be far cheaper than running and analyzing a new set of simulations from scratch[4], as information about the phase space is already known and all new regions can be explored on timescales shorter than the enhanced sampling runs. If the kinetics between states and equilibrium populations of states are changed by the perturbation, enhanced sampling is a promising method to more rapidly sample mutant systems. Enhanced sampling on a slow coordinate that is conserved between system conditions may be able to give unbiased exploration along faster coordinates, accelerating simulation of the entire phase space across conditions. This type of study of a system is complementary to post-simulation analysis methods that combine information from multiple states.

Enhanced sampling methods aim to use prior information about a system to accelerate simulation. In Metadynamics[5–9], a commonly-used enhanced sampling method and the focus of this work, time-dependent Gaussians are deposited along user-selected collective variables (CVs). This biases the system away from regions of phase space that have already been visited. However, the selection of which CV to use is critical for meaningfully sampling the system. A choice of poor CVs, even in the simplest of cases, leads to hysteresis such that the timescales required for convergence approach, or even exceed, unbiased sampling timescales[10].

To address this problem of CV choice, we recently showed[3,11] that time-structure based independent component analysis (tICA), a relatively recent advance in the Markov state model (MSM) field[12–14], yields an excellent set of linear and non-linear collective variables (CVs) for enhanced sampling via Metadynamics or other methods. tICA and its many variants[12,15,16] attempt to linearly approximate the dominant eigenfunctions of the Markovian Transfer operator[2,17]. We also showed that the tICA modes, also called tICs, can be well approximated in the low data regime even if *no* transition was observed in the unbiased training simulation[11]. We argued[11] that using the eigenfunctions of the transfer operator represent a natural basis for enhanced sampling since they approximate the slowest dynamical modes. Biasing a metadynamics on a tICA mode still represents a user-selected CV, but is intended to be the most slowly-decorrelating CV, thereby hopefully maximizing exploration of phase space.

Traditional tICA analyses produce linear models, which limit their descriptive capabilities. In contrast, non-linear tICA methods employ the popular kernel trick[12,15,18,19] which greatly increases their ability to approximate the Transfer operator. However, these kernel methods require the user to select both a distance metric and an appropriate kernel (i.e. Gaussian, exponential, polynomial). For computational efficiency, these kernel methods additionally require identification of appropriate landmark locations[11,15] and other parameters. Empirically, we have observed that a poor choice of landmarks and parameters can lead to slow convergence in sampling conformation space and their relative free-energies.

To introduce non-linearity into our CV without the drawbacks of kernel tICA, we turn towards deep neural networks (DNNs) for dimensionality reduction. The recent development of time-lagged extensions of variational auto-encoders, namely the Variational Dynamics Encoder (VDE) [20] and the Time-Lagged Auto-encoder (TAE)[21], have made the flexibility of deep neural networks available for dimensionality reduction in molecular simulation. Both VDEs and TAEs are variants of traditional auto-encoders, unsupervised machine learning algorithms that learn encodings for the input data by trying to reconstruct the high dimensional data from a low-dimensional encoded value. In contrast to the traditional method, the VDE and TAE frameworks attempt to reconstruct future dynamics (Figure 1a) based upon the current encoded value.

In this work, we show how the latent coordinate (z) of a VDE model can be used to perform enhanced sampling. This single latent layer encodes information regarding *all* of the slow modes of the system. A model trained on one protein can also be transferred to perform enhanced sampling on closely related mutants (Figure 1b). While previous papers have attempted to use traditional auto-encoders for enhanced sampling, these methods suffer from three flaws.

1). It is not immediately clear what the model "learns". For example, a traditional auto-encoder, similar to principal component analysis, might incorrectly identify fast floppy movement as being important for reconstruction. The black-box nature of the neural network only makes understanding what the model has learned more difficult. Our previous work with VDE protein saliency maps[20], and the tICA-VDE extension discussed in the FIP section below significantly improves our understanding of not just what the model is learning but also what atomic coordinates are being accelerated.

2). It is possible that auto-encoder-based models, given no information about dynamics, learn a representation that artificially adds barriers between "kinetically" similar states. As an example, consider a fictitious landscape with 3 basins (A-C) such that state A has to pass through state B to get to state C. It is possible that a traditional auto-encoder could map those states to non-continuous integers (A:1, B:3, C:2). This is a perfectly good dimensionality reduction but a bad coordinate for sampling via Metadynamics because a particle in basin A will now need to overcome additional repulsive forces when it jumps to the final basin C. By contrast, the VDE's loss function is designed to reproduce *time-lagged* dynamics which will naturally map A-C to continuous integers.

3). None of these methods show how these networks might be transferred to new, unseen mutants. In the present work, we show how a simple sequence alignment can be used to transfer these networks to new mutant proteins.

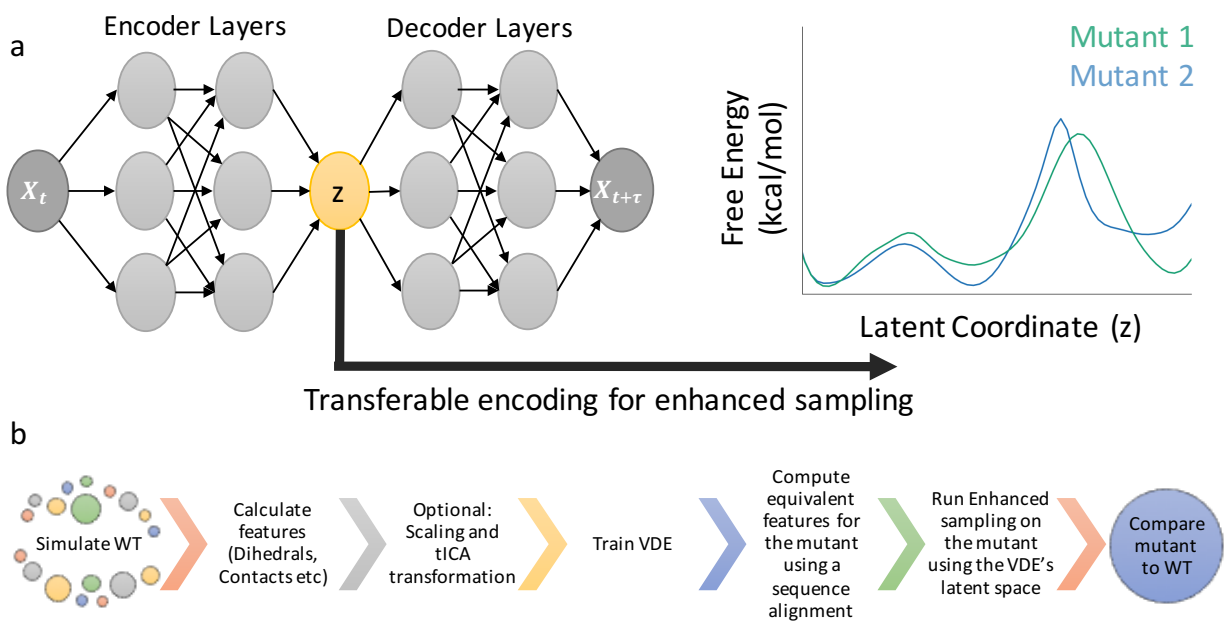

*Figure 1: Pictorial representation of the method. a). The VDE is a dimensionality reduction scheme designed to reproduce dynamics at a given lag time τ. The latent space z can then be used as a CV for enhanced sampling for not just the protein sequence that was used to train the model but also for related sequences. b). Step-wise algorithm for sampling a mutation after training a VDE on the protein wild type (WT).*

**VDEs create a single transferable encoding for alanine dipeptide.**

As a proof of concept, we train a VDE to learn the dynamics of alanine dipeptide (henceforth referred to as Alanine) and use the coordinates for enhanced sampling. We obtained a previously generated ~170ns solvated alanine dipeptide trajectory[3,11] (Figure 2b). This trajectory was run in the AMBER99sb[22] forcefield model (FF), and contained 2 transitions along the slower $\phi$ coordinate. The rarity of this transition makes thermodynamic analysis difficult, yet enhanced sampling allows us to obtain more converged statistics.

We first create a VDE with a latent coordinate that separates all major Alanine states ($\alpha_L, \alpha_R,$ and $\beta$) defined on its Ramachandran plot (Figure 2a). We generated the model using 170ns training data (Figure 2b). We then used the latent coordinate of the trained model for enhanced sampling in both the original FF of the training data, to obtain more converged sampling, and transferred to CHARMM22[23,24] to efficiently sample Alanine's landscape across multiple "mutant" FFs.

For the VDE model, we used the sin-cosine transform of the backbone dihedrals ($\phi, \psi$) as inputs for the neural network. For the encoder network, these four input values were fed into 2 fully connected layers with 16 hidden nodes with the Swish non-linearity[25] in the middle. This layer was then compressed to a single encoding (z), which was used for enhanced sampling calculations. Numerically, the architecture can be represented as 4-16-16-1, where the integers indicate the number of input, hidden layer, and latent nodes, respectively.

For the decoder network, the singular latent node was expanded using the same architecture as the encoder network but in reverse after passing the latent node through a variational ($\lambda$) noise layer[20] (1-$\lambda$-16-16-4). We trained the model using the dual time-lagged reconstruction and auto-correlation losses as suggested previously[20]. We trained the model at a lag-time of 1ns for 30 epochs. The training was performed using the Adam optimizer[26] with an initial learning rate of $1 \times 10^{-4}$. The model was built using PyTorch[27] and training took less than 5 minutes on a CPU platform. The trained model's results are shown in Figure 2 a-b. As can be seen in Figure 2a, the latent coordinate learns a highly non-linear transformation over the alanine dipeptide simulation, separating the major states along a single coordinate. In contrast, the slowest tICA solution primarily describes movement only along the slower $\phi$ coordinate. While the tICA coordinate can be used for sampling[3,11], an additional coordinate is needed to distinguish basins along $\psi$. Alternatively, simulations utilizing enhanced sampling on the first tICA coordinate would need to be run for long enough that the simulations naturally equilibrate along the faster $\psi$ coordinate. In contrast, a single VDE coordinate distinguishes between all 3 major alanine basins and forms the collective variable for our simulations.

To perform enhanced sampling simulations of Alanine using the latent VDE coordinate, we translated the fitted PyTorch VDE network into custom Plumed[28] expressions and performed the simulations using OpenMM[1]. At its core, the VDE encoding (z) is simply a non-linear combination of input dihedral features optimized using the dual loss function mentioned above. Once this collective variable has been transferred, sampling along this coordinate can be accelerated using a variety of CV-based enhanced sampling methods, such as Metadynamics, Adaptive Bias Force Sampling, Umbrella Sampling, etc.

The results of our well-tempered Metadynamics[29] simulations are shown in Figure 2c-d. Supporting Information Table 1 contains the simulation parameters, though we empirically found that a range of parameter values gave similar results. In our simulations, we observed fast diffusion along the VDE coordinate in less than 40 ns and obtained multiple transitions along the slower $\phi$ coordinate. Lastly, the VDE-Metadynamics simulations can be re-weighted to full phase space using MBAR[30] and last-bias reweighting or the time-independent estimator method[9], allowing for arbitrary projections along other observables (Figure 2d).

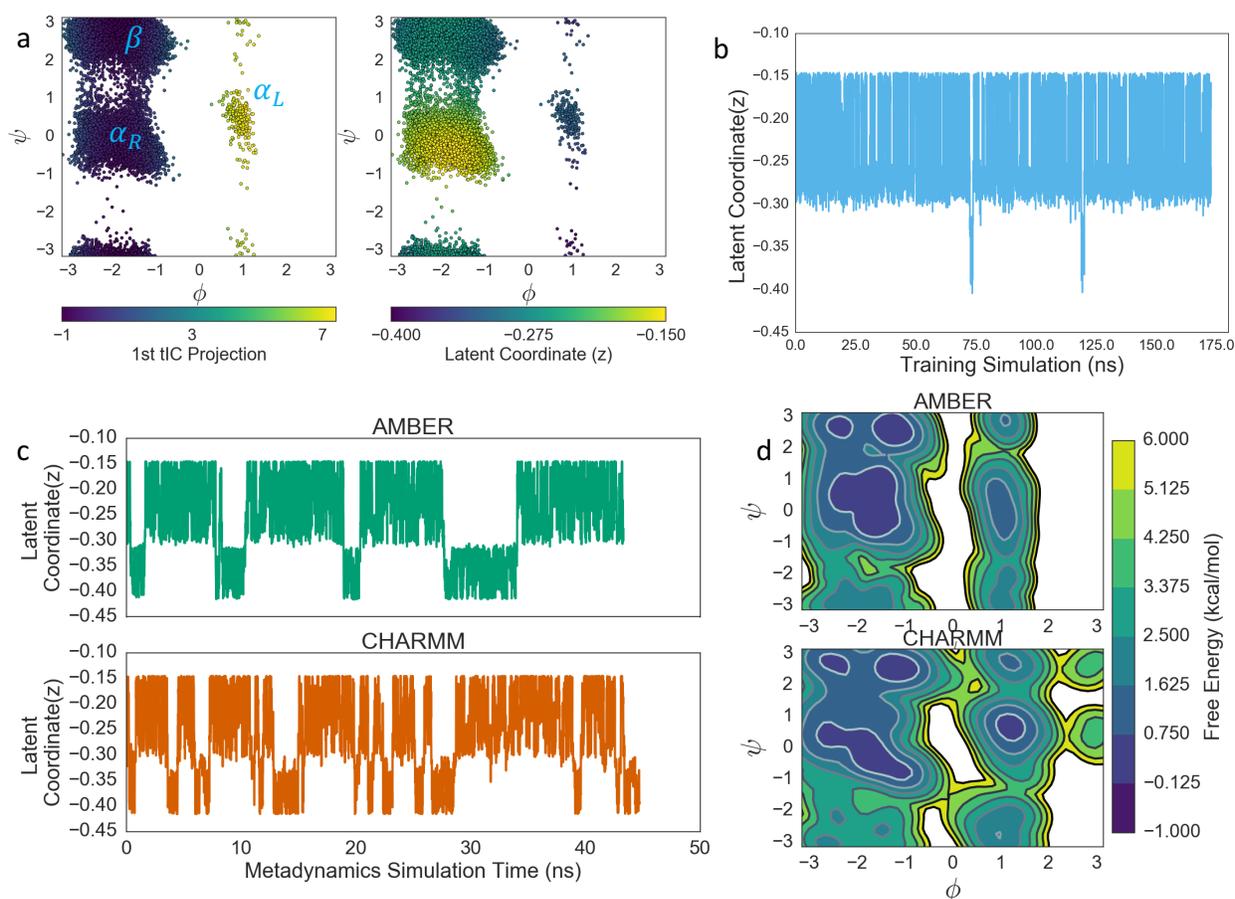

*Figure 2: Alanine dipeptide can be efficiently sampled using VDE-Metadynamics. a). Projection of the 170ns training simulation from Ref.[11] along 1) the slowest tICA coordinate and 2) the VDE latent coordinate. The VDE coordinate captures all major alanine basins in a single coordinate. b). Projection of the Alanine trajectory as a function of simulation time shows two rare transitions, making it difficult to compute thermodynamic quantities. c). Projection of the WT-Metadynamics results in both Amber and CHARMM forcefields using the transferable VDE network. d). The VDE-Metadynamics simulations can be re-weighted to full phase space using MBAR[30,31] which can then be projected onto the Ramachandran plot using existing libraries[32].*

## For larger proteins, neural network training can be accelerated by encoding the outputs of a linear tICA model

Despite the success of our VDE enhanced sampling framework demonstrated above, training directly on input features is unlikely to be efficiently scalable to large systems with many features. For instance, the commonly-used contact map featurization for a protein of length $n$ results in $O(n^2)$ features. For very small proteins, this is tractable; however, a protein with 30 residues already has 435 distance features. In neural network architectures, it is not uncommon to require the number of units in a hidden layer to be larger than the number of input features in order to capture nonlinear effects. Thus even in the first hidden layer, we would need to perform on the order of hundreds of thousands of float multiples (435 nodes * 435 multiples per node). These calculations will need to be performed at every single time step, thereby reducing simulation speed. Given this scaling problem, how could we tractably use the VDE on a larger system without limiting featurization schemes or hand-selecting features?

To address this scaling issue, we demonstrate that transforming the original feature input space to projections in tICA[12,13,15,33] space prior to encoding in the VDE alleviates this scaling problem (tICA-VDE). tICA seeks to embed protein configurations along their set of slowest de-correlating coordinates. We can project the data along a selected number of slowest tICA modes, also called tICs, such that the rest of the modes have an exchange timescale smaller than the length of our enhanced simulations, thereby ensuring convergence. Our collective variable *z* is now a non-linear combination of the current frame's projection onto the training dataset's slowest orthogonal tICA collective modes (Figure 1). Thus, our *single* hidden node encodes information about *all* the slow linear tICA modes of interest. While our methodology still requires the user to pick an appropriate feature space ( dihedrals/contacts/etc), we note that multiple feature schemes can be used simultaneously with an appropriate scaling term, and that the feature space can also be potentially be optimized using the WT simulation[34,35].

There are several key advantages of using tICA due to its ability to explicitly model the slowest modes[11,13] in the system. The tICA pre-processing step additionally allows us to train the networks much faster since the hidden node sizes can now be significantly lowered. This is because only a few tICA modes (< 5-10) are necessary to accurately capture the slow subspace of the system since the enhanced molecular simulation can naturally equilibrate across the remaining faster modes. The reduced network dimensionality also means that our model can actually be used as an efficient collective coordinate without excessively slowing down the simulations. Another advantage of using tICA is that if the protein dynamics are coupled i.e. moving along one tIC requires some change along the second (Figure 3a) , then we naturally include that coupling. It also allows us to understand[4] what the network is accelerating since we know what the tICs represent (Supporting Information Figure 3) at the atomic scale. An added advantage of this is that we can limit our simulations to only certain regions of phase-space, say to prevent sampling some irrelevant high free energy regions by simply excluding tICs that represent movement into that region. Other methods to sample multiple collective modes in parallel have been proposed, but tICA-VDE samples multiple collective modes while avoiding setting up a series of parallel bias or Hamiltonian replica exchange simulations[6,36] and the accompanying murky parameter selection.

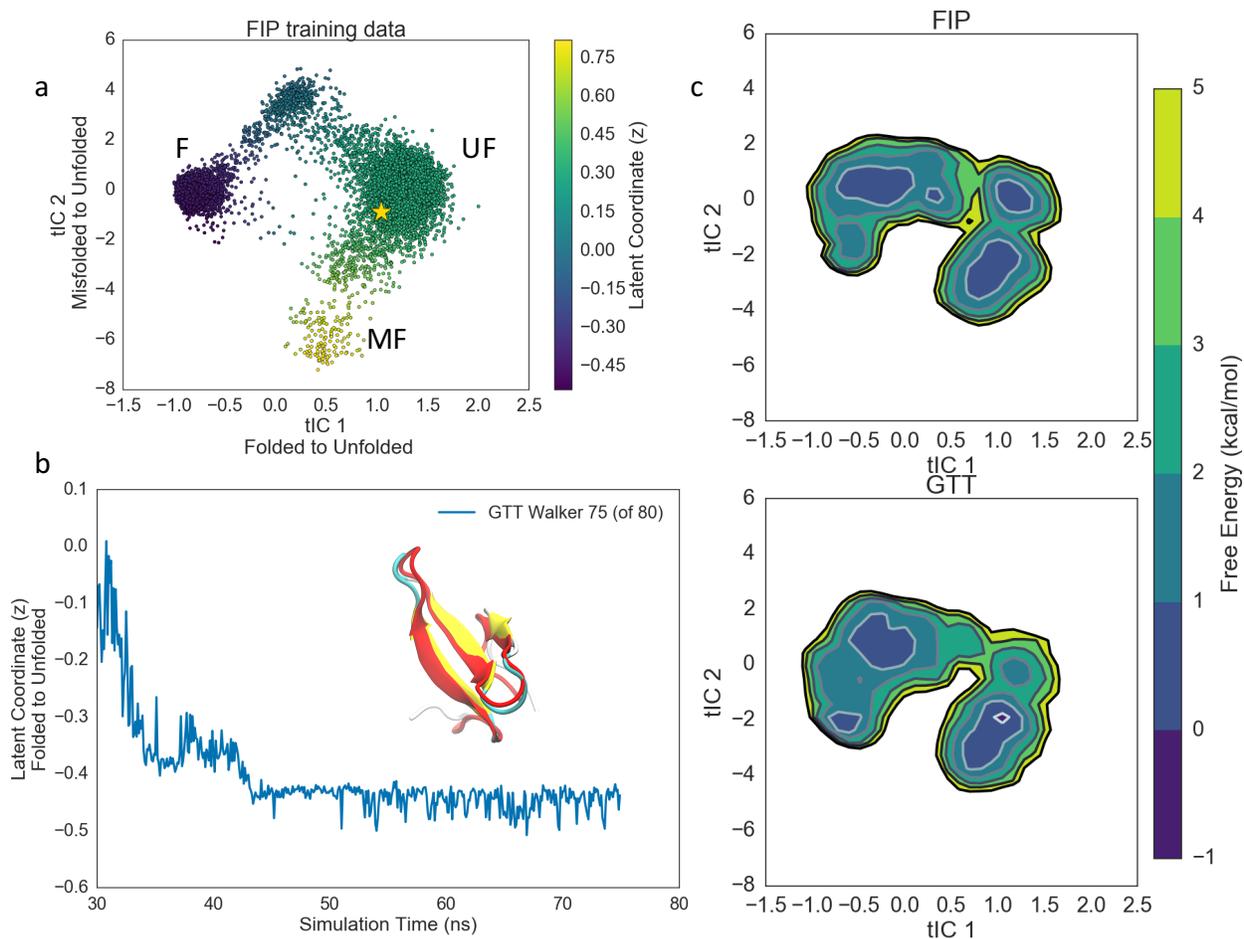

*Figure 3: VDE modeling of FIP creates a transferable coordinate for sampling mutants. a). The D.E. Shaw FIP[37,38] simulation data projected on to the slowest two tICA modes, tIC 1 and tIC2, show the presence of a folded (F), unfolded (UF), mis-folded (MF) states, similar to previous works[3]. We compress both of the slow modes onto a single coordinate (color bar) using a VDE model. The gold star indicates the starting point for all FIP and GTT simulations. b). The model is transferable to the GTT mutant of the FIP protein. The x-axis tracks the simulation time (the first 30ns were discarded for equilibration), the y-axis tracks projection along the VDE latent coordinate (z), and the inset shows that the predicted GTT folded state is very similar to the FIP folded state (folded state in color, predicted folded state in red). c). The Metadynamics simulations are clustered[39] and reweighted[9] using MBAR[30,31] to project on to the tICA coordinates. The 0 kcal/mol is defined using the highest populated FIP state. The protein images were generated using VMD[40,41] while the rest used IPython Notebook[42] and MSMExplorer[32].*

As a proof of concept, we recaptured the effects of the GTT mutation on the FIP WW domain folding landscape[43,38,37]. To that end, we obtained the $600\,\mu s$ FIP WW domain folding trajectories from the DE Shaw group[38,37] performed using the Anton machine[44].

We computed the closest heavy atom and alpha carbon contact distances for all residues at least 3 apart in sequence for the FIP training simulation for a total of 184 features. These were then mean-centered and scaled to unit variance. We used the Sparse-tICA[13,45] algorithm which performs the tICA calculation but with a tunable penalty for model complexity, allowing us to create a tICA model with a reduced number of features per tIC. This penalty is tunable: increasing the penalty reduces the number of features and results in fewer features needed to drive the system in enhanced sampling, but a too-sparse model leads to hidden orthogonal modes since we now have to wait for discarded degrees of freedom to respond to the bias. For our simulations, we selected a model that retained 106 of the 184 contact distance features across the top two tICs. Thus at every integration time step, SparsetICA saves us 78 contact distance calculations. Similarly to our previous work[3] and others[46,47], our tICA analysis indicated that the slowest two modes, both of which have an exchange timescale longer than 50 ns, corresponded to the folding process and an off-pathway register shift mis-folding process (Figure 3a, Supporting Information Figure 3). Our modeling also showed that we could approximate the tICA solutions with as little as 20-30% of the 600 $\mu s$ long trajectories (Supporting Information Figure 4), indicating that our methodology can likely scale to data-poor regimes as well. In previous work, no linear coordinate was able to distinguish between the folded, unfolded, and mis-folded states, thereby previously requiring[3] multiple coordinates to be simultaneously enhanced and connected via Hamiltonian replica exchange. To more simply perform enhanced sampling on both orthogonal modes, we trained a VDE model on the FIP simulation data projected onto the top two sparse tICs.

The tICA-transformed data was fed into a VDE network with an encoder architecture of 2-20-20-1. The decoder mirrored the encoder architecture except for an additional $\lambda$-layer. The VDE model was trained for 100 epochs using the Adam optimizer and an initial learning rate of 1 x $10^{-2}$. This entire pipeline was then transferred to Plumed[28] for enhanced sampling via well-tempered Metadynamics[29]. As shown in Figure 3a, the VDE latent coordinate (color bar) is able to transform the linear tICA modes into a highly non-linear function which goes from the folded (F) to unfolded (UF) to register shifted misfolded (MF) states. Without the preprocessing, a VDE transformation is unable to fully distinguish the folding process and misfolded state Thus, the latent coordinate of the VDE trained on tICA-transformed data was used as a transferable collective variable for Metadynamics simulations.

All of our simulations were set up as previously described[3], and the exact parameters for the Metadynamics simulations are given in SI Table 2. We started 80 walkers for both the FIP and GTT mutant from the same initial coordinates (the GTT mutant was homology modeled into the same conformation as FIP). Each walker was run for approximately 100ns, with the first 30ns being discarded for equilibration. For both FIP and GTT, we obtained ~ 7.5 $\mu s$ of aggregate sampling, but it is worth noting that all of these simulations were completed in only 4 days on commodity K80 GPUs. We also did not optimize the Metadynamics parameters, but believe that this sampling can be significantly accelerated by either selecting better parameters, such as initial Gaussian height, Gaussian drop rate or bias factor, or coupling to a structural reservoir of homology modeled states[3].

The results of our simulations are shown in Figure 3b-c. For both FIP and the GTT mutant, several walkers (4 for FIP, 11 for GTT) naturally folded to their WW topology (Figure 3b inset, Supporting Information Movie 1 and Movie 2) without providing any additional information about the folded states. Moreover, other walkers sampled the mis-folded state (Figure 3c), indicating the ability of the single non-linear coordinate to separate all major basins in the WW domain.

Lastly, we also clustered the FIP Metadynamics simulation to 5 states using the MiniBatch KMeans algorithm, and obtained macrostate populations after reweighting via MBAR[30,31]. Projecting onto either the tICA coordinates (Figure 3c) or the Latent coordinate (Supporting Information Figure 1) showed that the folded GTT state is slightly more stable (<1 kcal/mol) compared to FIP. In their paper[38], Piana et al. also showed that a triple mutation (the GTT mutant) stabilized the folded state by less than 1 kcal/mol. However, inherent robustness issues with Metadynamics simulations combined with limited methods for post-error analysis of Metadynamics simulations make it difficult to make precise predictions. However, we did repeat our Metadynamics simulations with different parameters (Supporting Information table 3 and Supporting Information Figure 2) and obtained qualitatively similar results in only ~4-5 μs of aggregate sampling per mutant. We believe it might be possible to push the aggregate simulation time down even further, but that optimization is outside the scope of this paper.

**Discussion and Conclusions**
In this work, we have shown how a new extension of traditional auto-encoders, namely the variational dynamics encoder (VDE), can provide an excellent *single* collective variable (CV) for enhanced sampling of protein dynamics. For simple systems that are inherently low dimensional, like the alanine dipeptide, one can learn the dynamics directly from the few degrees of freedom. One benefit of this approach is a more direct connection to those original degrees of freedom. However, for larger systems, it is more natural to employ some dimensionality reduction scheme, which creates another layer for interpretation, but allows for very efficient learning. Therefore we also demonstrate that pre-processing the training trajectories using the tICA[13] algorithm allows for more efficient training to produce simpler collective variables that directly push the system along *all* of its relevant slow modes. The flexibility and scalability of our method to larger systems allows us to use a single collective variable to rapidly perform enhanced sampling on related protein mutants.

While the purpose of this paper was to create a single transferable collective variable using deep neural networks, there are several extensions possible to this framework. One possibility would be to create an end-to-end training procedure, for instance by using a convolutional neural network, which constrains the dimensions at every hidden layer, so that the tICA transformation becomes unnecessary without sacrificing simulation speed.

While our method was used to compress multiple orthogonal processes a single collective variable, one can readily imagine more topologically complex free energy landscapes that do not make sense to be compressed to a single CV. In this case, our method could also be used with a higher-dimensional latent variable, which are sampled using the bias-exchange or parallel bias metadynamics methods mentioned earlier[6,36]. Another trivial extension would be to use the tICA algorithm or other optimization protocols as a "feature selection" tool so that it choses what features are fed into the network. This could potentially make better VDE models that are still small enough that they can be efficiently sampled. From an engineering perspective, our methodology of writing custom Plumed scripts is almost certainty inefficient and would likely benefit from some ability to embed these networks directly into the MD engine.

Our enhanced sampling scheme also presents an opportunity to engineer network architectures better suited for molecular simulations. For example, are dynamics better represented in "fatter" networks with more nodes and relatively few layers, or "deeper" networks with more layers but fewer nodes? Is there a difference in performance between different non-linear transformations (Sigmoid, ELU, ReLU, leaky ReLU etc.)? Can we better describe what these networks are learning at the atomic scale? For all of these tests, we recommend the Shaw FIP and GTT datasets[43,38,37] as a standard benchmark: to our knowledge, both the WT and mutant GTT domains contain at least 2 slow modes and being able to efficiently sample them remains challenging. Furthermore, several computational and experimental studies predict an observable difference between the two mutants.

Our current manuscript and previous works[3,11] draw inspiration from transfer learning, a commonly-used method in machine learning where a model trained on one dataset shows utility for a related dataset[48]. Similarly, we believe neural networks or other models built using the WT datasets can be transferred to mutant simulations to enhance dynamics, potentially allowing for MD to be more predictive. There are many biophysical parameters that determine protein conformation and that would be useful to be able to rapidly vary and characterize via simulation. Our method could enable performing sets of simulations for a protein with different mutations, post-translational modifications, ionic concentrations, protonation states, solvents, and so forth. In addition to a transfer of collective variable, other information from the WT model might be useful for setting Metadynamics parameters (such as Gaussian heights, sigma or the well-tempered bias factor)[3] or even accelerating convergence by coupling to structural reservoirs[3].

Ultimately, it is not known when transfer learning will fail for efficient sampling of related systems. Moffet et al.[49] argue that answer is likely system dependent. It is worth noting that any defined observable quantity from a molecular simulation is a collective variable and guaranteed to converge conditioned on enough sampling. However, a poorly-chosen CV could require more sampling to converge than brute force MD. Therefore, similar to previous work[3], we recommend caution against arbitrarily transferring these networks.

We believe auto-encoder frameworks, such as the VDE and TAE models, offer a promising path forward for enhancing dynamics, transferring models within related systems, and ultimately allowing for probing larger and more complex biophysical systems via simulation and making simulations predictive against experimental data.

**Acknowledgments**

MMS acknowledges support from NSF-MCB-0954714. HKWS acknowledges support from the NSF GFRP. This work used the XStream computational resource, supported by the National Science Foundation Major Research Instrumentation program (ACI-1429830). The authors would like to acknowledge Keri McKiernan for her suggestion to use tICA as a pre-processing step for the network. The authors would like to thank D.E. Shaw and DESERES for graciously providing the FIP folding trajectories. The authors also thank Brooke E. Husic for a critical reading of the manuscript.

**Software and Data Availability:**
All the code needed to regenerate the models and the Plumed input files presented in this paper are available freely online at www.github.com/msultan/vde_metadynamics. The code is licensed under the MIT license. The generated FIP and GTT trajectories are available upon request.

**Videos:**
Due to upload size limits, the videos accompanying this article have been put online instead.
https://www.youtube.com/watch?v=vEdumvjLEUc
https://www.youtube.com/watch?v=QSz6ML8zWbg

(1) Eastman, P.; Swails, J.; Chodera, J. D.; McGibbon, R. T.; Zhao, Y.; Beauchamp, K. A.; Wang, L. P.; Simmonett, A. C.; Harrigan, M. P.; Stern, C. D.; Wiewiora, R. P.; Brooks, B. R.; Pande, V. S. OpenMM 7: Rapid Development of High Performance Algorithms for Molecular Dynamics. *PLoS Comput. Biol.* **2017**, *13*, e1005659.
(2) Bowman, G. R.; Pande, V. S.; Noé, F. *An Introduction to Markov State Models and Their Application to Long Timescale Molecular Simulation*; 2014; Vol. 797.
(3) Sultan, M. M.; Pande, V. S. Transfer Learning from Markov Models Leads to Efficient Sampling of Related Systems. *J. Phys. Chem. B* **2017**, acs.jpcb.7b06896.
(4) Sultan, M. M.; Denny, R. A.; Unwalla, R.; Lovering, F.; Pande, V. S. Millisecond Dynamics of BTK Reveal Kinome-Wide Conformational Plasticity within the Apo Kinase Domain. *Sci. Rep.* **2017**, *7*, 15604.
(5) Laio, A.; Gervasio, F. L. Metadynamics: A Method to Simulate Rare Events and Reconstruct the Free Energy in Biophysics, Chemistry and Material Science. *Reports Prog. Phys.* **2008**, *71*, 126601.
(6) Pfaendtner, J.; Bonomi, M. Efficient Sampling of High-Dimensional Free-Energy Landscapes with Parallel Bias Metadynamics. *J. Chem. Theory Comput.* **2015**, *11*, 5062–5067.
(7) Sun, R.; Dama, J. F.; Tan, J. S.; Rose, J. P.; Voth, G. A. Transition-Tempered Metadynamics Is a Promising Tool for Studying the Permeation of Drug-like Molecules through Membranes. *J. Chem. Theory Comput.* **2016**, *12*, 5157–5169.
(8) Abrams, C.; Bussi, G. Enhanced Sampling in Molecular Dynamics Using Metadynamics, Replica-Exchange, and Temperature-Acceleration. *Entropy* **2013**, *16*, 163–199.
(9) Tiwary, P.; Parrinello, M. A Time-Independent Free Energy Estimator for Metadynamics. *J. Phys. Chem. B* **2015**, *119*, 736–742.
(10) Pan, A. C.; Weinreich, T. M.; Shan, Y.; Scarpazza, D. P.; Shaw, D. E. Assessing the Accuracy of Two Enhanced Sampling Methods Using Egfr Kinase Transition Pathways: The Influence of Collective Variable Choice. *J. Chem. Theory Comput.* **2014**, *10*, 2860–2865.
(11) Sultan, M. M.; Pande, V. S. TICA-Metadynamics: Accelerating Metadynamics by Using Kinetically Selected Collective Variables. *J. Chem. Theory Comput.* **2017**, *13*, 2440–2447.
(12) Schwantes, C. R.; Pande, V. S. Modeling Molecular Kinetics with tICA and the Kernel Trick. *J. Chem. Theory Comput.* **2015**, *11*, 600–608.
(13) Schwantes, C. R.; Pande, V. S. Improvements in Markov State Model Construction Reveal Many Non-Native Interactions in the Folding of NTL9. *J. Chem. Theory Comput.* **2013**, *9*,

2000–2009.
(14) Pérez-Hernández, G.; Paul, F.; Giorgino, T.; De Fabritiis, G.; Noé, F.; Perez-hernandez, G.; Paul, F. Identification of Slow Molecular Order Parameters for Markov Model Construction. *J. Chem. Phys.* **2013**, *139*, 15102.
(15) Harrigan, M. P.; Pande, V. S. Landmark Kernel tICA For Conformational Dynamics. *bioRxiv* **2017**.
(16) McGibbon, R. T.; Pande, V. S. Identification of Simple Reaction Coordinates from Complex Dynamics. **2016**, 1–16.
(17) Nüske, F.; Keller, B. G.; Pérez-Hernández, G.; Mey, A. S. J. S.; Noé, F. Variational Approach to Molecular Kinetics. *J. Chem. Theory Comput.* **2014**, *10*, 1739–1752.
(18) Hofmann, T.; Schölkopf, B.; Smola, A. J. Kernel Methods in Machine Learning. *Annals of Statistics*, 2008, *36*, 1171–1220.
(19) Müller, K. R.; Mika, S.; Rätsch, G.; Tsuda, K.; Schölkopf, B. An Introduction to Kernel-Based Learning Algorithms. *IEEE Trans. Neural Netw.* **2001**, *12*, 181–201.
(20) Hernández, C. X.; Wayment-Steele, H. K.; Sultan, M. M.; Husic, B. E.; Pande, V. S. Variational Encoding of Complex Dynamics. **2017**.
(21) Wehmeyer, C.; Noé, F. Time-Lagged Autoencoders: Deep Learning of Slow Collective Variables for Molecular Kinetics. **2017**.
(22) Lindorff-Larsen, K.; Piana, S.; Palmo, K.; Maragakis, P.; Klepeis, J. L.; Dror, R. O.; Shaw, D. E. Improved Side-Chain Torsion Potentials for the Amber ff99SB Protein Force Field. *Proteins* **2010**, *78*, 1950–1958.
(23) MacKerell, A. D.; Bashford, D.; Bellott, M.; Dunbrack, R. L.; Evanseck, J. D.; Field, M. J.; Fischer, S.; Gao, J.; Guo, H.; Ha, S.; Joseph-McCarthy, D.; Kuchnir, L.; Kuczera, K.; K Lau, F. T.; Mattos, C.; Michnick, S.; Ngo, T.; Nguyen, D. T.; Prodhom, B.; Reiher, W. E.; Roux, B.; Schlenkrich, M.; Smith, J. C.; Stote, R.; Straub, J.; Watanabe, M.; Wiórkiewicz-Kuczera, J.; Yin, D.; Karplus, M. All-Atom Empirical Potential for Molecular Modeling and Dynamics Studies of Proteins.
(24) Best, R. B.; Zhu, X.; Shim, J.; Lopes, P. E. M.; Mittal, J.; Feig, M.; MacKerell, A. D. Optimization of the Additive CHARMM All-Atom Protein Force Field Targeting Improved Sampling of the Backbone Φ, ψ and Side-Chain $\chi_1$ and $\chi_2$ Dihedral Angles. *J. Chem. Theory Comput.* **2012**, *8*, 3257–3273.
(25) Ramachandran, P.; Zoph, B.; Le, Q. V. Searching for Activation Functions. **2017**.
(26) Kingma, D. P.; Ba, J. Adam: A Method for Stochastic Optimization. **2014**.
(27) Paszke, A.; Chanan, G.; Lin, Z.; Gross, S.; Yang, E.; Antiga, L.; Devito, Z. Automatic Differentiation in PyTorch. **2017**, 1–4.
(28) Tribello, G. A.; Bonomi, M.; Branduardi, D.; Camilloni, C.; Bussi, G. PLUMED 2: New Feathers for an Old Bird. *Comput. Phys. Commun.* **2014**, *185*, 604–613.
(29) Barducci, A.; Bussi, G.; Parrinello, M. Well-Tempered Metadynamics: A Smoothly Converging and Tunable Free-Energy Method. *Phys. Rev. Lett.* **2008**, *100*.
(30) Shirts, M. R.; Chodera, J. D. Statistically Optimal Analysis of Samples from Multiple Equilibrium States. *J. Chem. Phys.* **2008**, *129*, 124105.
(31) Scherer, M. K.; Trendelkamp-Schroer, B.; Paul, F.; Pérez-Hernández, G.; Hoffmann, M.; Plattner, N.; Wehmeyer, C.; Prinz, J. H.; Noé, F. PyEMMA 2: A Software Package for Estimation, Validation, and Analysis of Markov Models. *J. Chem. Theory Comput.* **2015**,

*11*, 5525–5542.

(32) Hernández, C. X.; Harrigan, M. P.; Sultan, M. M.; Pande, V. S. MSMExplorer: Data Visualizations for Biomolecular Dynamics. *J. Open Source Softw.* **2017**, *2*.

(33) Noé, F.; Clementi, C. Kinetic Distance and Kinetic Maps from Molecular Dynamics Simulation. *J. Chem. Theory Comput.* **2015**, *11*, 5002–5011.

(34) McGibbon, R. T.; Pande, V. S. Variational Cross-Validation of Slow Dynamical Modes in Molecular Kinetics. *J. Chem. phyics* **2015**, *142*.

(35) Husic, B. E.; McGibbon, R. T.; Sultan, M. M.; Pande, V. S. Optimized Parameter Selection Reveals Trends in Markov State Models for Protein Folding. *J. Chem. Phys.* **2016**, *145*, 194103.

(36) Piana, S.; Laio, A. A Bias-Exchange Approach to Protein Folding. *J. Phys. Chem. B* **2007**, *111*, 4553–4559.

(37) Shaw, D.; Maragakis, P.; Lindorff-Larsen, K. Atomic-Level Characterization of the Structural Dynamics of Proteins. *Science (80-. ).* **2010**, *330*, 341–347.

(38) Piana, S.; Sarkar, K.; Lindorff-Larsen, K.; Guo, M.; Gruebele, M.; Shaw, D. E. Computational Design and Experimental Testing of the Fastest-Folding β-Sheet Protein. *J. Mol. Biol.* **2011**, *405*, 43–48.

(39) Harrigan, M. P.; Sultan, M. M.; Hernández, C. X.; Husic, B. E.; Eastman, P.; Schwantes, C. R.; Beauchamp, K. A.; Mcgibbon, R. T.; Pande, V. S. MSMBuilder: Statistical Models for Biomolecular Dynamics. *Biophys. J.* **2016**, *112*, 10–15.

(40) Humphrey, W.; Dalke, A.; Schulten, K. VMD: Visual Molecular Dynamics. *J. Mol. Graph.* **1996**, *14*, 33–38, 27–28.

(41) Frishman, D.; Argos, P. Knowledge-Based Protein Secondary Structure Assignment. *Proteins* **1995**, *23*, 566–579.

(42) Pérez, F.; Granger, B. E. IPython: A System for Interactive Scientific Computing. *Comput. Sci. Eng.* **2007**, *9*, 21–29.

(43) Lindorff-Larsen, K.; Piana, S.; Dror, R. O.; Shaw, D. E. How Fast-Folding Proteins Fold. **2011**, *517*.

(44) Shaw, D. E.; Chao, J. C.; Eastwood, M. P.; Gagliardo, J.; Grossman, J. P.; Ho, C. R.; Ierardi, D. J.; Kolossváry, I.; Klepeis, J. L.; Layman, T.; McLeavey, C.; Deneroff, M. M.; Moraes, M. A.; Mueller, R.; Priest, E. C.; Shan, Y.; Spengler, J.; Theobald, M.; Towles, B.; Wang, S. C.; Dror, R. O.; Kuskin, J. S.; Larson, R. H.; Salmon, J. K.; Young, C.; Batson, B.; Bowers, K. J. Anton, a Special-Purpose Machine for Molecular Dynamics Simulation. In *Proceedings of the 34th annual international symposium on Computer architecture - ISCA '07*; ACM Press: New York, New York, USA, 2007; Vol. 35, p. 1.

(45) McGibbon, R. T.; Husic, B. E.; Pande, V. S. Identification of Simple Reaction Coordinates from Complex Dynamics. *J. Chem. Phys.* **2016**, *146*, 1–16.

(46) Krivov, S. V. The Free Energy Landscape Analysis of Protein (FIP35) Folding Dynamics. *J. Phys. Chem. B* **2011**, *115*, 12315–12324.

(47) Wan, H.; Zhou, G.; Voelz, V. A. A Maximum-Caliber Approach to Predicting Perturbed Folding Kinetics Due to Mutations. *J. Chem. Theory Comput.* **2016**, *12*, 5768–5776.

(48) Torrey, L.; Shavlik, J. Transfer Learning. In *Handbook of Research on Machine Learning Applications*; IGI Global, 2009; pp. 242–264.

(49) Moffett, A. S.; Shukla, D. On the Transferability of Time-Lagged Independent

Components between Similar Molecular Dynamics Systems. **2017**.

# Supporting Information for Transferable neural networks for enhanced sampling of protein dynamics

Mohammad M. Sultan[1], Hannah K. Wayment-Steele[1], & Vijay S. Pande[1†]

[1]Department of Chemistry, Stanford University, 318 Campus Drive, Stanford, California 94305, USA.

[†]pande@stanford.edu

| Parameter | Value |
| --- | --- |
| **Gaussian Height** | 1 kj/mol |
| **Gaussian width** | 0.01 |
| **Bias Factor** | 6 |
| **Gaussian Drop rate** | 2ps |
| **Feature Space** | Dihedrals |
| **Normalized Features** | True |
| **Sim. Save rate** | 10ps |
| **Sim. Temp** | 300K |

Table 1: Set of parameters used for the VDE -Metadynamics simulations of Alanine dipeptide across 2 different FFs

| Parameter | Value |
| --- | --- |
| **Gaussian Height** | 0.5 kj/mol |
| **Gaussian width** | 0.05 |
| **Bias Factor** | 20 |
| **Gaussian Drop rate** | 2ps |
| **Feature Space** | Ca contacts + Closest heavy atom contacts |
| **Normalized Features** | True |
| **tICA transformed** | True (Sparse Tica) |
| **tICA lag time** | 25 ns |
| **tICs** | 2 |
| **Grid** | -.75, 1.07 (Based off the training data +30% on either side) |

| Interval | -0.58,0.8 (Based off the training data) |
|---|---|
| **Sim. Save rate** | 100 ps |
| **Sim. Temp** | 395K |
| **Number of walkers** | 80 |
| **Walker read stride** | 10000 |

Table 2: Set of parameters used for the VDE -Metadynamics simulations of both the FIP and GTT WW domains

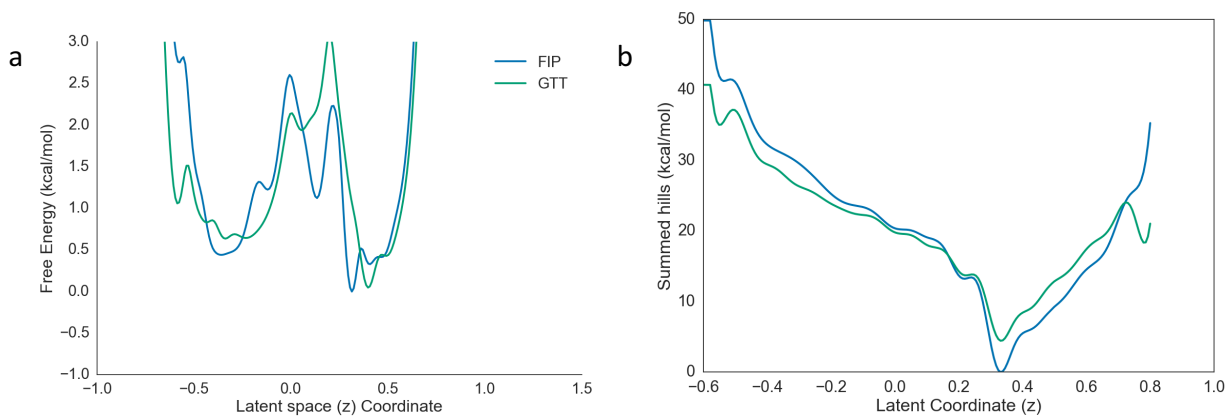

Figure 4: MBAR reweighted projection of the FIP and GTT simulation onto the Latent space shows that the folded state (<-.45) becomes slightly more stable upon the GTT mutation. b). Similar results are obtained by simply summing the hills using Plumed to get an estimate of the deposited bias. In both cases, 0kcal/mol is the lowest value in the FIP ensemble.

| Parameter | Value |
|---|---|
| **Gaussian Height** | 0.1 kj/mol |
| **Gaussian width** | 0.05 |
| **Bias Factor** | 100 |
| **Gaussian Drop rate** | 5ps |
| **Sim. Save rate** | 100 ps |
| **Sim. Temp** | 395K |
| **Number of walkers** | 60 |
| **Walker read stride** | 10000 |

Table 3: Set of parameters used in the replication study for the VDE -Metadynamics simulations of both the FIP and GTT WW domains. In this case, we wanted to see the effects of adding a smaller bias for longer. The total sampling in this case was 5 $\mu s$ (2-3 $\mu s$ less than in Table 2). Again, the first 30ns of each walker was discarded for equilibration. The results are given in the figure below.

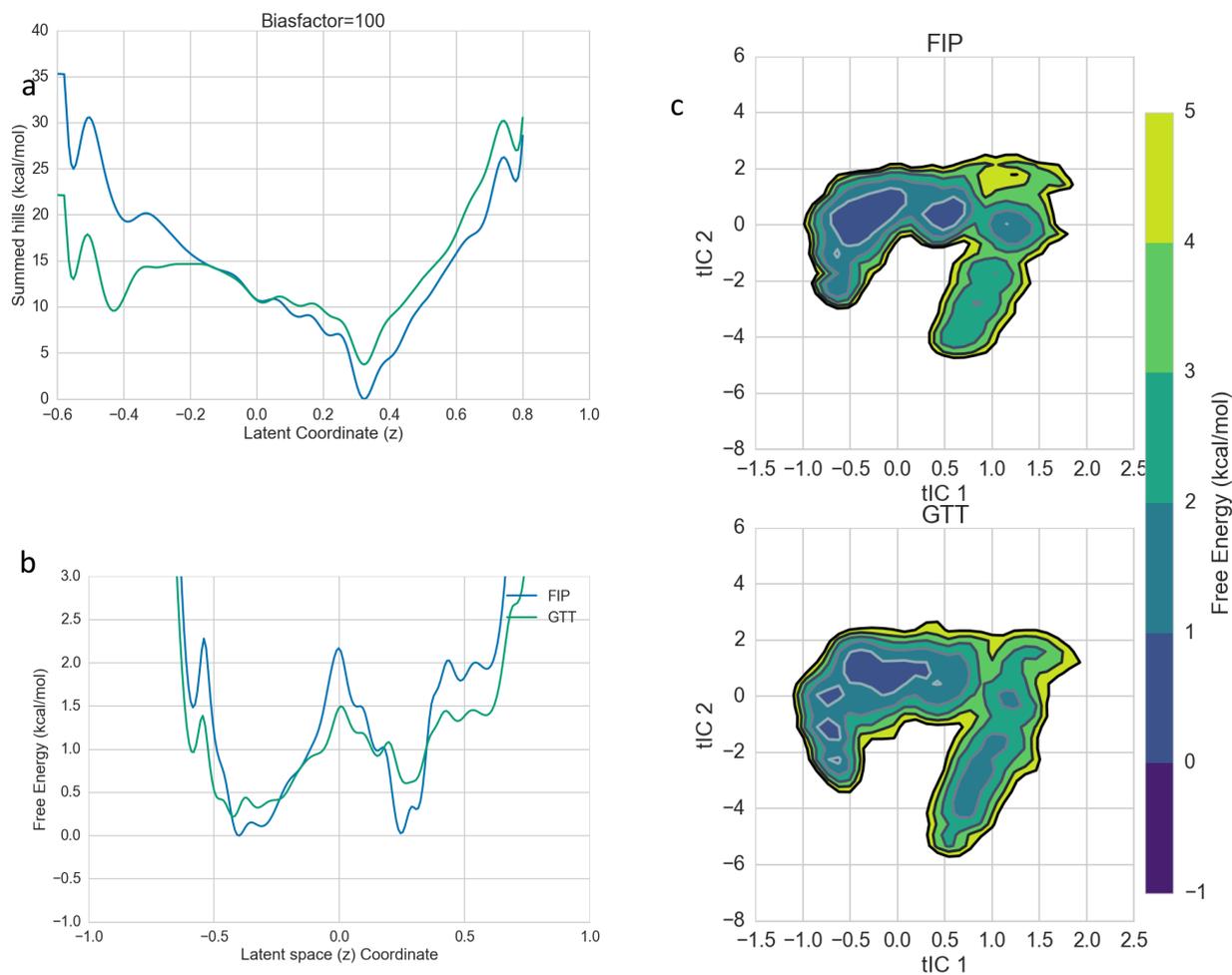

Figure 5: Results for the replication study using different Metadynamics parameters (Table 2 above) give similar result. a). Integrated bias showing that the GTT folded state (z<-.4) is more stable than FIPs. b-c) Similar results are obtained after coarse graining the frames to 5 states and using MBAR to re-weight the dynamics.

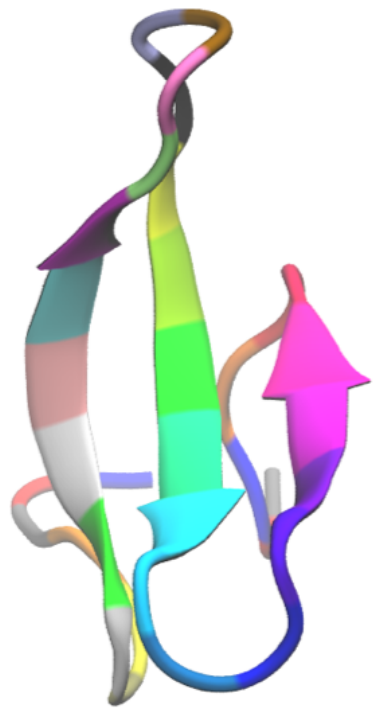 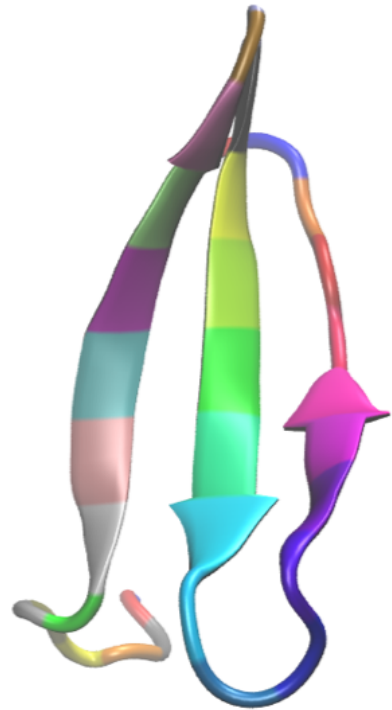

FIP FOLDED STATE                GTT MIS-FOLDED STATE

*Figure 6: Comparison of the GTT misfolded state to the FIP folded state reveals the presensce of a register shift in the first two beta sheets. For example, the residue colored salmon in both is hydrogen bonding with residue green in FIP and light green/teal in the register shifted GTT state.*

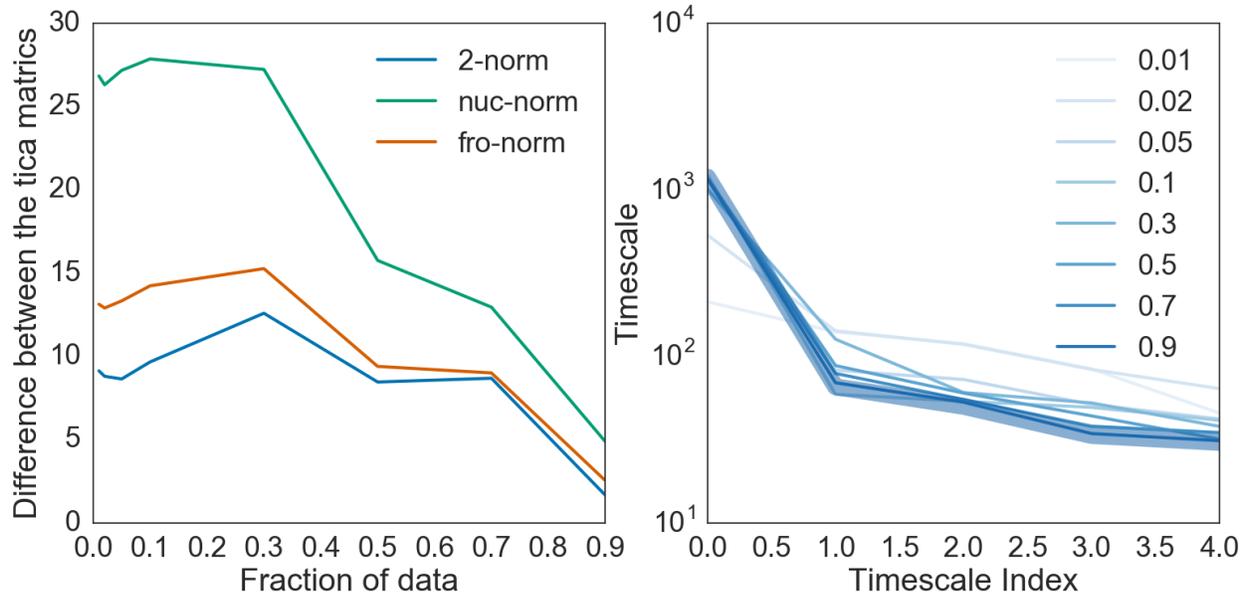

Figure 7: Time dependence test of convergence of the tica solution and timescales as a function of trajectory length. For each trajectory, we limited it to somewhere between 1 and 90% of its final length and recomputed the Sparse tica model. We then compared both the tICA matrices (5tics *184 features) and the longest timescale against the final 600 µs solution. As it can be seen, at about 20-30%, both the tica matrices and the timescales start to converge to their final values. The thick bar on the right represents the solution with 100% of the data.